\documentclass[fleqn,usenatbib]{mnras}

\usepackage[T1]{fontenc}
\usepackage{ae,aecompl}
 
\usepackage{xcolor}
\usepackage{url}
\usepackage{footnote}

\usepackage{graphicx}	
\usepackage{amsmath}	
\usepackage{tablefootnote}

\usepackage{amssymb}	
\usepackage{siunitx}
\usepackage{pdflscape}

\usepackage{enumitem}
\setlist{leftmargin=7mm}

\title[DESOM real-bogus classifier]{Self-Supervised Clustering on Image-Subtracted Data with Deep-Embedded Self-Organizing Map}

\author[Mong et al.]{Y. -L. Mong,$^{1,2}$\thanks{E-mail: yik.mong@monash.edu}
K. Ackley,$^{1,2,3}$
T. L. Killestein$^{3}$,
D. K. Galloway$^{1,2,4}$,
M. Dyer$^{5}$,
\newauthor
R. Cutter$^{3}$,
M. J. I. Brown$^{1}$,
J. Lyman$^{3}$,
K. Ulaczyk$^{3}$,
D. Steeghs$^{3}$,
\newauthor
V. Dhillon$^{5,12}$,
P. O'Brien$^{6}$,
G. Ramsay$^{7}$,
K. Noysena$^{8}$,
R. Kotak$^{9}$,
\newauthor
R. Breton$^{10}$,
L. Nuttall$^{11}$,
E. Pall\'{e}$^{12}$,
D. Pollacco$^{3}$,
E. Thrane$^{1,2}$,
\newauthor
S. Awiphan$^{8}$,
U. Burhanudin$^{5}$,
P. Chote$^{3}$,
A. Chrimes$^{3}$,
E. Daw$^{5}$,
\newauthor
C. Duffy$^{7}$,
R. Eyles-Ferris$^{6}$,
B. P. Gompertz$^{3}$,
T. Heikkil\"{a}$^{9}$,
P. Irawati$^{8}$,
\newauthor
M. Kennedy$^{10}$,
A. Levan$^{3}$,
S. Littlefair$^{5}$,
L. Makrygianni$^{5}$,
T. Marsh$^{3}$,
\newauthor
D. Mata S\'{a}nchez$^{12,13}$,
S. Mattila$^{9}$,
J. R. Maund$^{5}$,
J. McCormac$^{3}$,
D. Mkrtichian$^{8}$,
\newauthor
J. Mullaney$^{5}$,
E. Rol$^{1,2}$,
U. Sawangwit$^{8}$,
E. Stanway$^{3}$,
R. Starling$^{6}$,
\newauthor
P. Str{\o}m$^{3}$,
S. Tooke$^{6}$,
K. Wiersema$^{3}$
\\ \\
$^{1}$School of Physics \& Astronomy, Monash University, Clayton VIC 3800, Australia\\
$^{2}$OzGrav: The ARC Centre of Excellence for Gravitational Wave Discovery, Clayton VIC 3800, Australia\\
$^{3}$Department of Physics, University of Warwick, Coventry, West Midlands, CV4 7AL, UK\\
$^{4}$Institute for Globally Distributed Open Research and Education (IGDORE) \\
$^{5}$Department of Physics and Astronomy, University of Sheffield, Sheffield, S3 7RH, UK\\
$^{6}$School of Physics and Astronomy, University of Leicester, University Road, Leicester, LE1 7RH, UK\\
$^{7}$Armagh Observatory \& Planetarium, College Hill, Armagh, BT61 9DB, Co.Armagh, Northern Ireland, UK\\
$^{8}$National Astronomical Research Institute of Thailand,  260  Moo 4, T. Donkaew,  A. Maerim, Chiangmai, 50180, Thailand\\
$^{9}$Department of Physics and Astronomy, University of Turku, FI-20014 Turun yliopisto, Finland\\
$^{10}$Department of Physics and Astronomy, University of Manchester, M13 9PL, UK\\
$^{11}$Institute of Cosmology and Gravitation, University of Portsmouth, Dennis Sciama Building, Burnaby Road, Portsmouth, PO1 3FX, UK\\
$^{12}$Instituto de Astrof\'{i}sica de Canarias, E-38205 La Laguna, Tenerife, Spain \\
$^{13}$Departamento de Astrof\'{i}sica, Univ. de La Laguna, E-38206 La Laguna, Tenerife, Spain
}

\date{Accepted XXX. Received YYY; in original form ZZZ}

\pubyear{2021}

\begin{document}
\newcommand{\Msun}{$M_{\odot}$}
\newcommand{\Lsun}{$L_{\odot}$}
\newcommand{\Rsun}{$R_{\odot}$}
\newcommand{\solar}{${\odot}$}
\newcommand{\comment}[1]{\textcolor{red}{{\texttt{#1}} }}

\label{firstpage}
\pagerange{\pageref{firstpage}--\pageref{lastpage}}
\maketitle

\begin{abstract}
Developing an effective automatic classifier to separate genuine sources from artifacts is essential for transient follow-ups in wide-field optical surveys. The identification of transient detections from the subtraction artifacts after the image differencing proccess is a key step in such classifiers, known as real-bogus classification problem.
We apply a self-supervised machine learning model, the deep-embedded self-organizing map (DESOM) to this ``real-bogus'' classification problem. DESOM combines an autoencoder and a self-organizing map to perform clustering in order to distinguish between real and bogus detections, based on their dimensionality-reduced representations. We use $32\times32$ normalized detection thumbnails as the input of DESOM. We demonstrate different model training approaches, and find that our best DESOM classifier shows a missed detection rate of $6.6\%$ with a false positive rate of $1.5\%$. DESOM offers a more nuanced way to fine-tune the decision boundary identifying likely real detections when used in combination with other types of classifiers, for example built on neural networks or decision trees. We also discuss other potential usages of DESOM and its limitations.
\end{abstract}



\section{Introduction}\label{sec:introduction}
Time-domain astronomy has risen in popularity among the astronomical research fields during the past decade. It is particularly important for the study of gamma-ray bursts \citep[GRBs;][]{piran04,zfd06,llt12,bfc13,jcd13,tlf13,ber14,kz15,cup15,kkl17,ltl19,caa20,hpb20,ack21,mag21} and gravitational-wave events \citep[GW;][]{aaa16,aaa17b,bbf17,cbk17,cbv17,cfk17,gvb17,hcm17,mbf17,sfk17,gcs20}. These events require prompt (timescales of hours) follow-up observations in order to identify their nature, before they become too faint to be detected \citep[see e.g.][]{rkl09}. %
Information about the event's origin can often be relatively poor, with uncertainty regions of hundreds of square degrees typical.
To this aim, facilities with a large field of view (FoV) enable the follow-up observations of these transient events.

The Gravitational-wave Optical Transient Observer (GOTO\footnote{\url{http://goto-observatory.org}}) is a robotic optical telescope designed to search for the  counterparts of GW events \citep{dsg20,sga21}. GOTO presently consists of two telescope arrays, each equipped with $8\times40\,{\rm cm}$ unit telescopes (UTs) and a total FoV of $\sim80$ square degrees per pointing. Other than following up GRBs and GW events, GOTO also performs regular all-sky survey observations in order to explore the transient sky with serendipitous searching. It can reach depths of $\approx20\,{\rm mag}$ in the adopted Baader $L$ band filter ($400$--$700\,{\rm nm}$), with a set of $3\times90\,{\rm seconds}$ exposures. The image size of each camera is $8176\times6132\,{\rm pixels}$ with a pixel scale of $\approx1.2\,{\rm arcsec}$.

Difference imaging is commonly used to identify transient objects on an image \citep{al98,bec15}. With a reference image, taken during a historical visit of the same field as the input image (also called the science image), the two images can be aligned by performing an affine transformation (e.g. with a custom Python package {\tt spalipy}). The aligned reference frame is then subtracted from the science frame (e.g. using {\tt HOTPANTS}) in order to generate a difference image \citep{bec15}. An ideal subtraction helps to remove the vast majority of the objects that do not vary in intensity. Transients which appear only on the science frame should appear on the difference image after the image subtraction process.

``Real-bogus'' classification is the process of separating real objects from ``bogus'' detections, including instrumental artifacts, subtraction residuals, bad pixels, etc, on the difference image \citep{brn12}. Due to imperfections in the difference imaging method, $\sim10^4$ subtraction artifacts can be identified as detections by {\tt SExtractor} per GOTO image \citep{ba96}. A large number of subtraction artifacts makes it impossible to manually separate real transients from bogus detections. As a result, an automatic real-bogus classifier is required. Supervised machine learning models are commonly used to construct the real-bogus classifier. Among all of the supervised learning algorithms, the convolutional neural network (CNN) model shows the best performance on solving computer vision problem \citep{cfe16, gbv17, crf17}. The current real-bogus classifier of GOTO is built based on the VGG16 CNN model \citep[hereafter GOTO-VGG;][]{sz14,dmm19,kls21}. 

There are two major disadvantages of using supervised machine learning to solve real-bogus classification problems. First, since the training process of the model is supervised, all training instances must be labelled as real or bogus. This labelling process requires a high cost of expert human labour. This challenge has been previously addressed by using detections on the science frame to train the classifier \citep{mag20} or building a training set with synthetic transients \citep{ssy20,kls21}. Second, during the actual prediction process, supervised learning models act like ``black box'' models. The algorithm is typically too complicated for a human to visually understand how the prediction is made. Therefore, improving the model is challenging.

Using unsupervised learning models to build our classifier, avoids the shortcomings of supervised learning models. Since the clustering model groups similar data together, it is easy to understand that two inputs with the same prediction means that they are close to each other in the parameter space, and hence likely arising from a similar origin.

In this paper, we apply unsupervised machine learning  to solve the real-bogus classification problem. In \S\ref{sec:intro_to_algorithms}, we introduce the learning algorithms used to construct our classifier. In \S\ref{sec:data_processing}, we discuss how to extract and pre-process our data before performing further analyses. In \S\ref{sec:analysis_results}, we discuss how we train our model and report the result of our best model. We also compare the model performance with the VGG16 model. In \S\ref{sec:discussion}, we discuss the advantages and the shortcomings of the model. Finally, we conclude our results in \S\ref{sec:conclusion}.

\section{Learning Algorithms} \label{sec:intro_to_algorithms}
We employ the deep-embedded self-organizing map \citep[DESOM;][]{fla19,tst21} as the unsupervised learning model to build our real-bogus classifier. DESOM consists of two parts, the autoencoder (AE) and the self-organizing map (SOM). In this section, we will be introducing the basic concept behind these algorithms.

\subsection{Autoencoder}
The autoencoder (AE) model is a variant of the neural network model, with the main objective of reconstructing the input data via dimensionality reduction \citep{bal12,whw14,bkg20}. The architecture of the AE is usually symmetric, i.e. the output has the same dimension as the input. Due to the objective of AE, the input $X$ is also used as the target in the training process. Therefore, AE is considered to be a ``self-supervised'' learning algorithm.

The AE architecture consists of three parts, the encoder, the ``bottleneck'' and the decoder. The bottleneck represents the middle layer of the AE. AE can generally be divided into two types, undercomplete and overcomplete. For the undercomplete AE, the number of neurons in the bottleneck layer is smaller than that of the input layer. On the other hand, an overcomplete AE has a bottleneck with more neurons than the input layer. In this work, we use undercomplete AE to construct our model in order to capture the most important features from the raw input data. 

The first half of the AE is defined as the encoder. It maps the input $X$ to the output parameter space, also called the ``latent space''. Since the bottleneck has a smaller size than the input layer, the output of the bottleneck is considered to be a compressed representation of the raw input. Therefore, the encoder can be used to perform dimensionality reduction. The second half of the AE is the decoder. It takes the compressed representation $\phi(X)$ as the input and attempts to reconstruct the original input $X$. The mathematical operation of an AE can be written as
\begin{align}
    \hat{X}=\psi(\phi(X)),
\end{align}
where $\phi$ and $\psi$ are the operators of the encoder and the decoder respectively. The reconstructed output of the AE is denoted by $\hat{X}$.

The encoder part of the AE is usually identical to an artificial neural network \citep[ANN;][]{dan13} or a convolutional neural network \citep[CNN;][]{on15} model. The choice of using the Convolutional ({\tt Conv2D}) layers or the fully-connected ({\tt dense}) layers depends on the type of the problem. The {\tt Conv2D} layer identifies local patterns under translation and rotation invariance. For computer vision problems, {\tt Conv2D} usually performs better in general. In this work, we use a CNN model to construct our AE model.

\subsection{Self-Organizing Map}
A Self-Organizing Map (SOM) is a clustering algorithm consisting of only two layers, the input layer and the output layer \citep{koh90,koh01}. Each layer consists of several nodes, and each input node is connected to all output nodes with a corresponding weight. Since those two layers are fully connected, SOM is a variant of an ANN model. The number of output nodes, which represents the number of desired clusters, is the most important hyperparameter of SOM.

Each cluster in the input parameter space is characterized by a weight vector, which is called the prototype vector (PV), of an output node. The SOM nodes usually form a two-dimensional grid, also called a SOM map. The input vector $\mathbf{x}$ is connected with an output node with the weights $\mathbf{w}_{ij}$, where $i,j$ indicates the position of the node in the output layer. Therefore, the PV of an $m\times m$ SOM map can be expressed as
\begin{align}
    \mathbf{W}=\begin{pmatrix}
    \mathbf{w}_{11} & \mathbf{w}_{12} & \cdots & \mathbf{w}_{1m} \\
    \mathbf{w}_{21} & \mathbf{w}_{22} & \cdots & \mathbf{w}_{2m} \\
    \vdots & \vdots & \ddots & \vdots \\
    \mathbf{w}_{m1} & \mathbf{w}_{m2} & \cdots & \mathbf{w}_{mm} \\
    \end{pmatrix}.
\end{align}
The prediction output of the SOM model is the cluster with the minimum Euclidean distance between $\mathbf{w}_{ij}$ and $\mathbf{x}$. We call the selected $\mathbf{w}_{ij}$ the ``winner PV'' of $\mathbf{x}$.

In the SOM training process, $\mathbf{W}$ is updated iteratively with three steps: competition, cooperation and adaptation. In the competition step, the input $\mathbf{x}$ searches for the winner PV $\mathbf{w_k}$ with the minimum Euclidean distance $|\mathbf{w_k}-\mathbf{x}|$ among $\mathbf{W}$. In the cooperation step, the distances between $\mathbf{w}_{\mathbf{k}}$ and other PVs $\mathbf{w}_{ij}$ are calculated. The winner PV $\mathbf{w_k}$ will then be updated by reducing the Euclidean distance between the $\mathbf{w_k}$ and the $\mathbf{x}$. However, $\mathbf{w_k}$ is not the only PV being updated. In fact, all $\mathbf{w}_{ij}$ will be updated in the final adaptation step depending on their distances from $\mathbf{w}_{\mathbf{k}}$. 

The update of any $\mathbf{w_i}$ is directional pointing towards $\mathbf{x}$ in the latent space, and the step size is controlled by three factors: the Euclidean distance between $\mathbf{w_k}$ and $\mathbf{x}$, the Euclidean distance between $\mathbf{w_k}$ and $\mathbf{w_i}$ and the learning rate $\eta$. The step size of the update also decays with the number of the training iterations $t$. Therefore, the algorithm of the update is written as
\begin{align}
    \mathbf{w_i}'=\mathbf{w_i}+\eta(t)h_{\mathbf{ik}}(t)|\mathbf{x}-\mathbf{w_k}|,\label{eq:som_update}
\end{align}
where
\begin{align}
    h_{\mathbf{ik}}(t)=\exp\left(-\frac{|\mathbf{w_i}-\mathbf{w_k}|^2}{T(t)^2}\right)
\end{align}
is the kernel function. We use the Gaussian neighborhood function to be the kernel here. The temperature parameter,
\begin{align}
    T(t)=T_0\exp\left(-\frac{t}{\tau_T}\right),\label{eq:temp}
\end{align}
determines the kernel size at the training epoch $t$. The maximum temperature and the decay constant are denoted by $T_0$ and $\tau_T$. The same form is used to describe the $t$-dependent learning rate,
\begin{align}
    \eta(t)=\eta_0\exp\left(-\frac{t}{\tau_\eta}\right).
\end{align}

\begin{figure*}
\centering
	\includegraphics[width=0.6\textwidth]{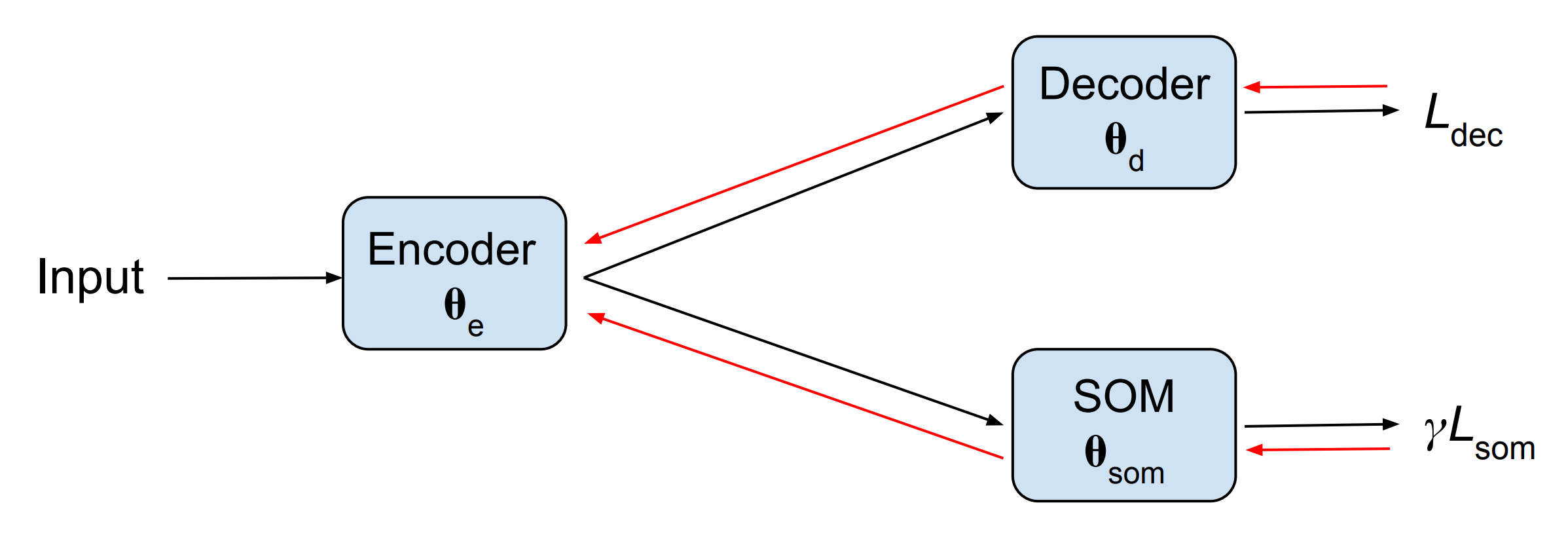}
    \caption{The propagation flows of DESOM training \citep{fla19}. The {\it black} lines indicate the direction of the forward propagation. And the {\it red} lines indicate the direction of the back-propagation.}
    \label{fig:propagation_flow}
\end{figure*}

\subsection{Deep-Embedded Self-Organizing Map}\label{sec:desom}
The Deep-Embedded Self-Organizing Map (DESOM) model is constructed by combining an AE and SOM. The AE achieves the dimensionality reduction, while the SOM performs the actual clustering process on the dimensionality-reduced input. Therefore, the SOM layer is attached to the bottleneck of the AE. With this model architecture, DESOM can be divided into two types depending on how the model is trained, the ``combine-trained'' DESOM, the ``separate-trained'' DESOM. 

\subsubsection{Combine-trained DESOM}\label{sec:combine_train_approach}
The combined training of DESOM was first demonstrated by \cite{fla19}. In the training process, the parameters in all layers are trainable. In the other words, both AE and SOM are trained at the same time. A combine-trained DESOM was also used to build an image-quality-based recommendation system \citep{tst21}. Figure \ref{fig:propagation_flow} illustrates the architecture of DESOM. As we can see that DESOM generates two separate outputs from the decoder and the SOM layer with their corresponding losses, the least square reconstruction loss $L_{\rm dec}$ and the SOM loss $L_{\rm som}$ indicating the Euclidean distance between the input $\mathbf{x}$ and the winning PV in the latent space. The total loss of each run can be defined as the weighted sum of the two losses,
\begin{align}
    L_{\rm tot}=L_{\rm dec}(\boldsymbol{\theta}_{\rm e},\boldsymbol{\theta}_{\rm d})+\gamma L_{\rm som}(\boldsymbol{\theta}_{\rm e},\boldsymbol{\theta}_{\rm som}),\label{eq:total_loss}
\end{align}
where $\gamma$ is the weight of the SOM loss. The trainable model parameters of the encoder, the decoder and the SOM are denoted by $\boldsymbol{\theta}_{\rm e}$, $\boldsymbol{\theta}_{\rm d}$ and $\boldsymbol{\theta}_{\rm som}$. All parameters connected from the output layer to the input layer are updated in each training iteration through a numerical algorithm called back-propagation \citep{mun10}. However, since we have two output layers, there are two different paths in the back-propagation to update the parameters (see the {\it red} lines in Figure \ref{fig:propagation_flow}). Therefore, the back-propagations of both decoder and SOM also contribute to the update of the encoder, and this is the reason why both losses also depend on $\boldsymbol{\theta}_{\rm e}$. 

There are two major disadvantages of using this training approach. First, DESOM has many adjustable hyperparameters such as the number of layers, the number of neurons, the size of the SOM map, among others, which makes the model evaluation with respect to different hyperparameter setups very complicated. Blindly searching for the best hyperparameter set is very time consuming. Second, this approach provides relatively ineffective SOM training. At the beginning of the SOM training phase, the kernel size $\tau_T$ is large, such that the update of the SOM map is more global. As a result, the early training phase of SOM is more effective to roughly map the SOM map onto the latent space of AE. However, the AE is under-trained during the early training phase. If we train both SOM and AE together, the SOM layer will learn the unuseful information from the under-trained AE latent space effectively due to a large $T(t\sim0)$. On the other hand, once the AE has been well-trained, the SOM kernel $h_{\mathbf{ik}}(t\gg\tau_T)$ has converged to a small size. At that time, the training process would only fine-tune the SOM layer, which means that the final SOM is mainly trained on an under-trained AE. To solve this problem, we are motivated to explore the following alternative training approach.

\subsubsection{Separate-trained DESOM}
\label{subsubsec:separate}
Here we instead break down the DESOM training into two separate processes. First, we train the AE individually. Then, we freeze both $\boldsymbol{\theta}_{\rm e}$ and $\boldsymbol{\theta}_{\rm d}$ (by manually setting them to be ``untrainable'' parameters). Finally, we insert the SOM layer to the bottleneck of the AE and train for it with the frozen, trained AE. 

Training the AE and SOM individually can effectively speed up the hyperparameter searching process. Considering that AE and SOM have $M$ and $N$ hyperparameter configurations respectively, the grid searching finds the best combined configuration among $M\times N$ configurations. By separating the training processes of AE and SOM, we first search for the best AE among those $M$ configurations. Then, we implement the best AE to search for the best SOM configuration. Therefore, the number of trials reduces down to $M+N$, which significantly speeds up the grid searching process.

Since the SOM training process commences after obtaining the best AE configuration, the SOM layer is directly trained on the well-established latent space of the AE bottleneck. Unlike the combine-trained DESOM, which the SOM is trained on an unlearnt bottleneck of the AE during the early training phase, the training process of the SOM layer in the separate-trained DESOM is more effective. Using this approach, all the shortcomings of the combined training approach can be addressed.


\subsection{VGG16 Classifier}
To provide a direct comparison to the DESOM architecture presented in this work, we used the pre-trained CNN classifier currently being used in the live GOTO pipeline, details of which are presented in full in \cite{kls21}. This model is a 330,000 parameter, 8 layer deep CNN that was trained on around 400,000 labeled examples. The model inherits the broad structure of the VGG16 network of \cite{sz14} by utilising ``conv-conv-pool'' blocks, but is significantly downscaled owing both to the smaller scale and overall lower complexity of astronomical images, compared to the datasets used in the computer science literature.

\section{Data Extraction and Processing} \label{sec:data_processing}
We randomly gathered 719 stacked images taken by GOTO between 16 April 2021 and 12 June 2021 to perform analyses in this work. Each of these images includes science, reference and subtracted frames. All images are processed through the GOTO standard image processing pipeline \citep{sga21}.

\begin{figure}
\centering
	\includegraphics[width=0.35\textwidth]{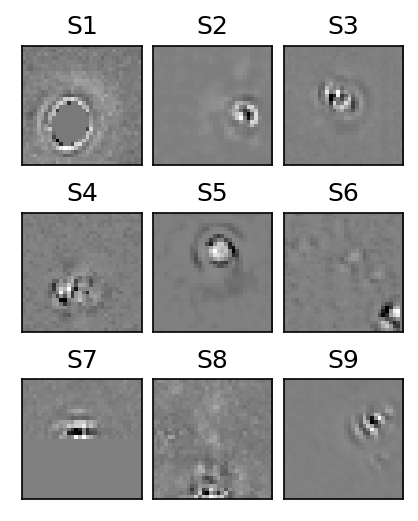}
    \caption{Examples of $32\times32$ pixel cutouts from GOTO prototype images extracted with the DC approach. S1 is an example of masked subtraction of a bright source. S6, S8 and S9 are likely due to the statistical fluctuation of the background. S7 is a detection lying on the edge of the difference image. These examples show that the DC approach usually does not center the subtraction residual within the thumbnail.}
    \label{fig:off_stamps}
\end{figure}

We use two different approaches, the Difference-Coordinate (DC) approach and the Science-Coordinate (SC) approach (see \S\ref{sec:dc_approach} and \ref{sec:sc_approach} for more details), to extract the coordinates of the detections from the difference images. Once we obtain the detection coordinates, we use a $32\times32$ pixel cutout centered on the coordinates to be the input thumbnail (see Figure \ref{fig:off_stamps} for some examples extracted with the DC approach). Since DESOM is an unsupervised learning model, labeling process is not required to build our training set.

\subsection{Difference-Coordinate Approach}\label{sec:dc_approach}
The DC approach is the usual approach of extracting the detection coordinates of candidate sources from the difference image. The coordinates are extracted by running the source extraction software {\tt SExtractor} on the difference images. We randomly extracted 1\,000\,000 detection cutouts from 719 difference images to build our data set (DC data set). We further split the data set into training and test sets, which contain 800\,000 and 200\,000 detections respectively.

\begin{figure}
\centering
	\includegraphics[width=0.4\textwidth]{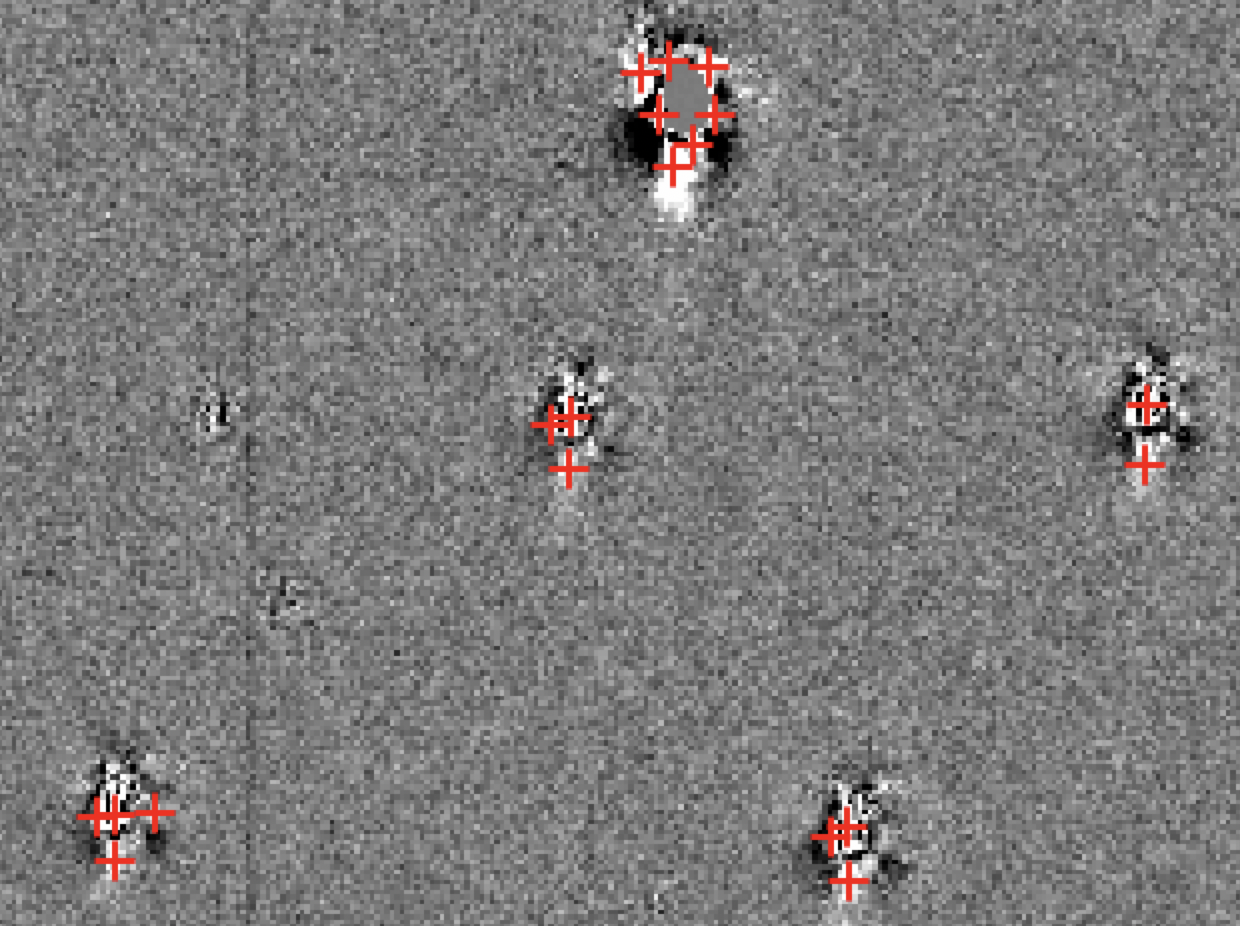}
    \caption{Example of the segmentation caused by difference imaging. The extracted detections recovered by {\tt SExtractor} are indicated with {\it red} crosses. The segmentation of the image subtraction can substantially increase the number of the bogus detections on a difference image.}
    \label{fig:segmentation}
\end{figure}

There are two obvious problems of using this approach to create inputs. First, as we can see in Figure \ref{fig:off_stamps}, the central points of most of the cutouts are offset from the original positions of the sources on the science images. This issue is caused by the fact that {\tt SExtractor} tends to identify the subtraction residuals surrounding the actual position of the source, as candidate detections. The root of this issue resides in the following {\tt SExtractor} performance method. During the subtraction process, pixel discretization and fractional shifts may result in point spread function (PSF) kernel mismatches between the science and reference frames. As a result, the science detections are segmented into multiple peaks extracted by {\tt SExtractor} on the difference image (see Figure \ref{fig:segmentation}). This effect substantially increases the number of bogus detections on the difference image. Thus, even a classifier with a low false positive rate (FPR) could result in many false positives. The disadvantages of this approach motivate us to develop another approach, the SC approach, to perform source extraction.

\subsection{Science-Coordinate Approach}\label{sec:sc_approach}
\begin{figure}
\centering
	\includegraphics[width=0.35\textwidth]{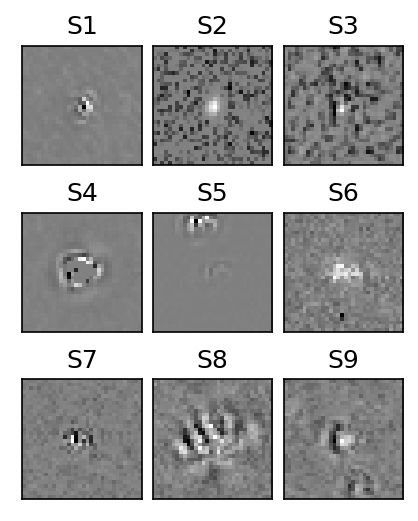}
    \caption{Examples of $32\times32$ pixel cutouts from GOTO prototype images extracted with the SC approach. Unlike the ones extracted with the DC approach (see Figure \ref{fig:off_stamps}), the subtraction residuals are centered at the middle of the thumbnails.}
    \label{fig:on_stamps}
\end{figure}

Here we run {\tt SExtractor} on the science frame instead of the difference frame, to extract the detection coordinates. Those coordinates will then be used to generate cutouts from the difference image. The thumbnail generated with the SC approach is centered at the original position of the source on the science image, as the difference image is astrometrically aligned to the science frame before the image subtraction (see Figure \ref{fig:on_stamps}). In addition, we can effectively eliminate the majority of the segmented detections which might arise from the subtraction of a single source. Therefore, the SC approach solves both problems of the DC approach.

Nevertheless, the SC approach has its own issues. Using the coordinates of all science detections to extract the cutouts is extremely inefficient, since the subtraction process can otherwise effectively reject  persistent objects. To solve this problem, we cross-match all detections on the science image with the ones on the difference image, and select only those science detections with a cross-matched result within a  radius defined in Eq.(\ref{eq:xmatch_r}). If the same detection appears on both science image and difference image, the offset in the cross-match result is defined as the ``subtraction offset''. However, it is very challenging to conclude that every detection on the science image is the same source of another detection on the difference image even if the subtraction offset is small. 

\begin{figure}
\centering
	\includegraphics[width=0.5\textwidth]{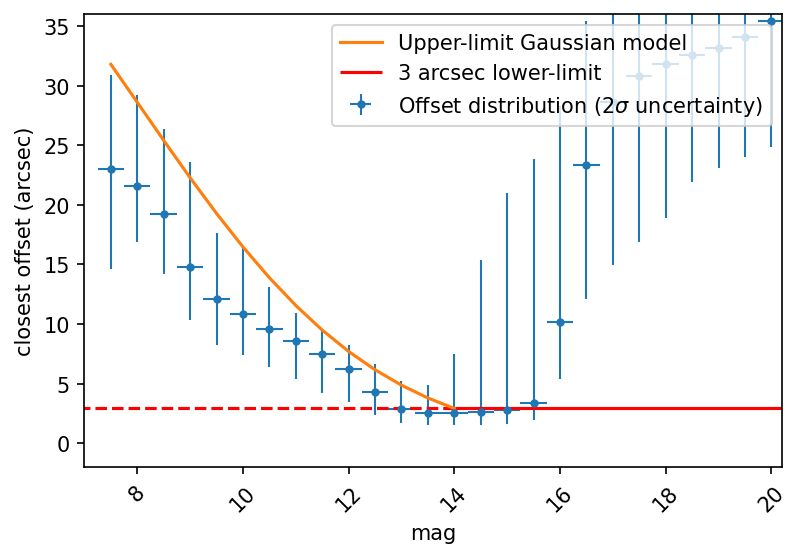}
    \caption{The offset between the detection on the science image and its closest detection on the difference image against its magnitude. The $y$-axis error-bar indicates the $2\sigma$ confidence level of the offset distribution. The {\it orange} curve represents the best Gaussian function fitting the $2\sigma$ upper-limits of the data with $m<13.5$. The {\it red} line indicates the lower-limit of $f(m)$ in Eq.(\ref{eq:xmatch_r}) at $3\,{\rm arcsec}$. For the science detection with $m>13.5$ and an offset smaller than $3\,{\rm arcsec}$, the closest detection on the difference image is considered as the same source.}
    \label{fig:off2mag}
\end{figure}

Although we can only conclude that a smaller offset implies a higher chance of association, we have to set a critical offset to claim the association of the detections on the science image and the difference image. We plot the offsets of the detections on the science images from the closest detections on difference images against the magnitudes of the detections on the science image (Figure \ref{fig:off2mag}). Depending on the brightness of the source, the closest offset could be a good representative of the subtraction offset. For the bright sources with $m\lesssim10$, the subtraction offset can go up to $\sim30\,{\rm arcsec}$ ($2\sigma$ confidence level), due to the effect of masking saturated pixels. For the fainter detections with $m\gtrsim15$, the offset gets larger with decreasing brightness. This can be explained by the fact that the subtraction on the faint stable source is usually cleaner. If the subtraction residual left on the difference image is negligible, it cannot be detected by {\tt SExtractor}. Therefore, in the cross-matching process, the science detection actually matches with another completely different source. Hence, the cross-matched offset becomes larger. We conclude that the closest offset beyond $m>13.5$ in Figure \ref{fig:off2mag} does not reflect the subtraction offset of the same source.

Since the subtraction offset is magnitude-dependent, we obtain the offset threshold as a function of magnitude by fitting a phenomenological piece-wise Gaussian function,
\begin{align}
    f(m)=\max\left[3, A\exp\left(-\frac{(m-m_0)^2}{2\sigma^2}\right)\right],\label{eq:xmatch_r}
\end{align}
to the data in Figure \ref{fig:off2mag}. We restrict our fitting at $2\sigma$ upper-limit of the subtraction offset and $m<13.5$. The best-fit parameters are $A=46.3$, $m_0=3.7$ and $\sigma=4.4$. We also set the lower limit of the offset threshold at $3\,{\rm arcsec}$. Any detection with magnitude $m$ on the science image which has a closest offset smaller than $f(m)$ is considered as having a cross-matched result, and it is included in the data set.

\begin{figure}
\centering
	\includegraphics[width=0.5\textwidth]{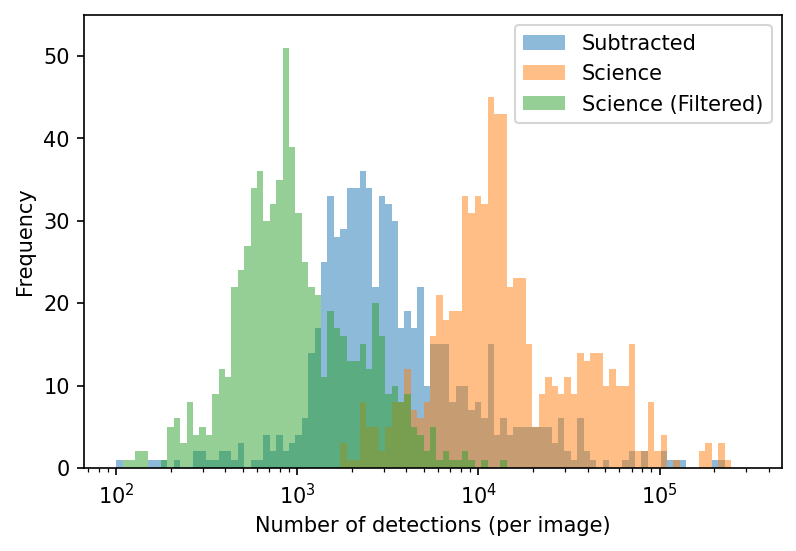}
    \caption{Distributions of numbers of science detections ({\it orange}), subtracted detections ({\it blue}) and extracted detections with the SC approach ({\it green}) per image. With the SC approach, the number of detections needed to be analyzed drops to $\sim10^{2-3}$ per image.}
    \label{fig:detnum}
\end{figure}

With this approach, the number of detections that need to be analyzed is greatly reduced. In Figure \ref{fig:detnum}, we can see that a science frame usually has $\sim10^4$ detections. After the difference imaging, only $\sim10^{3-4}$ detections are left on a difference image. With the SC approach, $\sim10^{2-3}$ detections per image have to be analyzed, which is an order of magnitude smaller than the original number of science detections. However, this filtering step has to be done carefully in order to prevent over-filtering. To evaluate how many detections are over-filtered, we estimate the fraction of the ``real-labeled'' detections which are not filtered in the SC approach. This estimation is done by manually inspecting 200 random detections from those over-filtered detections. We find that the SC approach can only recover $\approx80\%$ of those ``real-labeled'' detections. We also find that $\approx50\%$ of those over-filtered detections are only statistical fluctuations instead of genuine real detections. Therefore, the over-filtered rate of the SC approach is $\approx10\%$. For those over-filtered detections, most of them are filtered by mistake due to a slightly larger subtraction offset. This over-filtered rate is roughly consistent with the fact that we obtain our offset threshold $f(m)$ by fitting to the $2\sigma$ confidence level of the subtraction offset and set the lower limit at $3\,{\rm arcsec}$.

We further exclude all faint detections ($m>21$) and edge detections (lying within 50 pixels from the edge of the image) to obtain our second data set, the SC data set, which then contains 637\,937 detection thumbnails. This  set is further split into training and test sets, which contain 574\,143 and 63\,794 detections respectively.

\begin{figure}
\centering
	\includegraphics[width=0.3\textwidth]{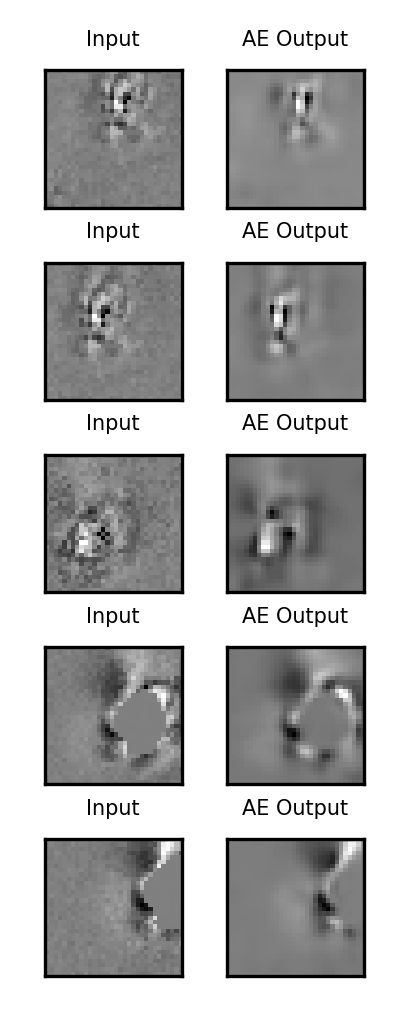}
    \caption{Examples of some cutout inputs from GOTO prototype images and their reconstructed outputs generated by the DC-trained AE. These examples show that the AE performs well on denoising with preserving most of the characteristics from the original inputs.}
    \label{fig:off_ae_reconstruct}
\end{figure}

\subsection{Normalization}
Image normalization is an important data pre-processing procedure to make sure that all inputs are consistent and all pixels have the same range of values. Due to different observing conditions, detections having the same intrinsic brightness but observed at different epochs usually have different background levels and peak values of the signals. Since the real-bogus classification is usually based only on the morphology of the detection on the difference image, we need to normalize the input (the pixel values of the subtracted cutouts) in order to avoid the classification being affected by these variations. 

We normalize each input by multiplying each difference cutout pixel value by
\begin{align}
    f(p)=\left\{
    \begin{array}{ll}
        [(p-\Bar{p})/(p_{1.00}-\Bar{p})+1], & p>\Bar{p} \\
        \left[\max\left(-|\Bar{p}-p|/|\Bar{p}-p_{0.05}|, -1\right)+1\right]/2, & p<\Bar{p},
    \end{array}
    \right.
\end{align}
where $\Bar{p}$ and $p_{1.00}$ are the median pixel value and the peak value of the thumbnail respectively.
$p_{0.05}$ is the 5-percentile value of the pixels which lie below $\Bar{p}$. In this normalization algorithm, we set $\Bar{p}$ to be the background level with a normalized value of 0.5. We then linearly normalize the above-background pixels ($p>\Bar{p}$) and below-background pixels ($p<\Bar{p}$) with different normalization constants. We normalize the peak value $p_{1.00}$ to be 1 and the 5-percentile $p_{0.05}$ to be 0. Since the subtraction process might generate some outliers with extremely negative values, in order to avoid the normalization scale being affected by those outliers, we use $p_{0.05}$ instead of the minimum pixel value to normalize the below-background pixels.

\subsection{Minor Planet (MP) Test Set}
To generate a test set for the classifiers we use the code of \cite{kls21} to extract a set of stamps centered on known minor planets. Positions of minor planets are queried using \texttt{SkyBoT} \citep{bvt06} and cross-matched to difference image sources to yield a confirmed set of genuine examples. We extract stamps of size $32\times32$ pixels centered on the difference image detections (the DC approach described in \S\ref{sec:dc_approach}), equivalent to extracting stamps centered on science sources given the lack of underlying template source. This approach yields 50\,279 real detections, spanning a wide range of sky conditions, PSFs, and source magnitudes.

We also randomly gathered 92\,901 human-reviewed bogus detections from the GOTO detection database to form a bogus test set. Combining with the 50\,279 real detections, the entire test set contains 143\,180 detections.

\section{Analyses and Results} \label{sec:analysis_results}
\subsection{Model Training} \label{sec:model_train}
We start our analysis by training our DESOM model with the DC data set (see \S\ref{sec:dc_approach}). We apply different training approaches (see \S\ref{sec:desom}) and perform grid searching on the hyperparameter space of the DESOM model to obtain the best model configuration. 

\begin{figure}
\centering
	\includegraphics[width=0.3\textwidth]{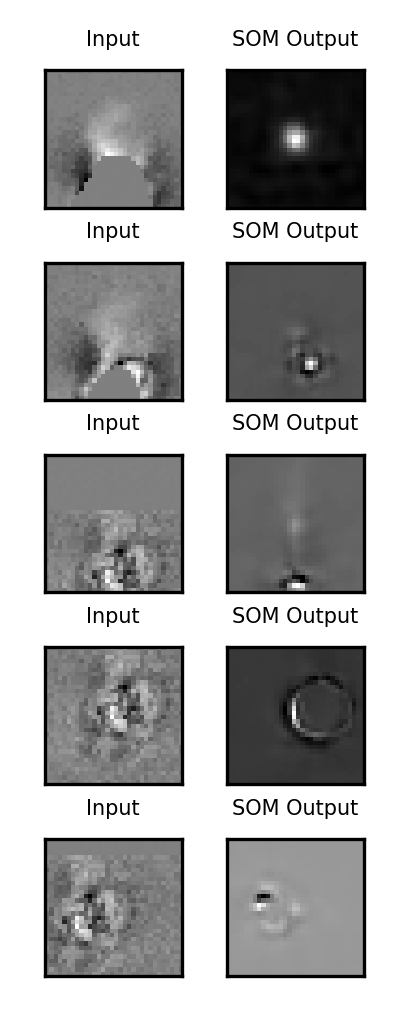}
    \caption{Examples of some cutout inputs and their decoded PVs showing the poor performance of the DC-trained DESOM. In the last three examples, the SOM layer classifies the same pattern located at different positions into three different PVs, which indicates that the prediction of SOM is not transformation invariant. }
    \label{fig:off_desom_reconstruct}
\end{figure}

\begin{figure*}
\centering
	\includegraphics[width=\textwidth]{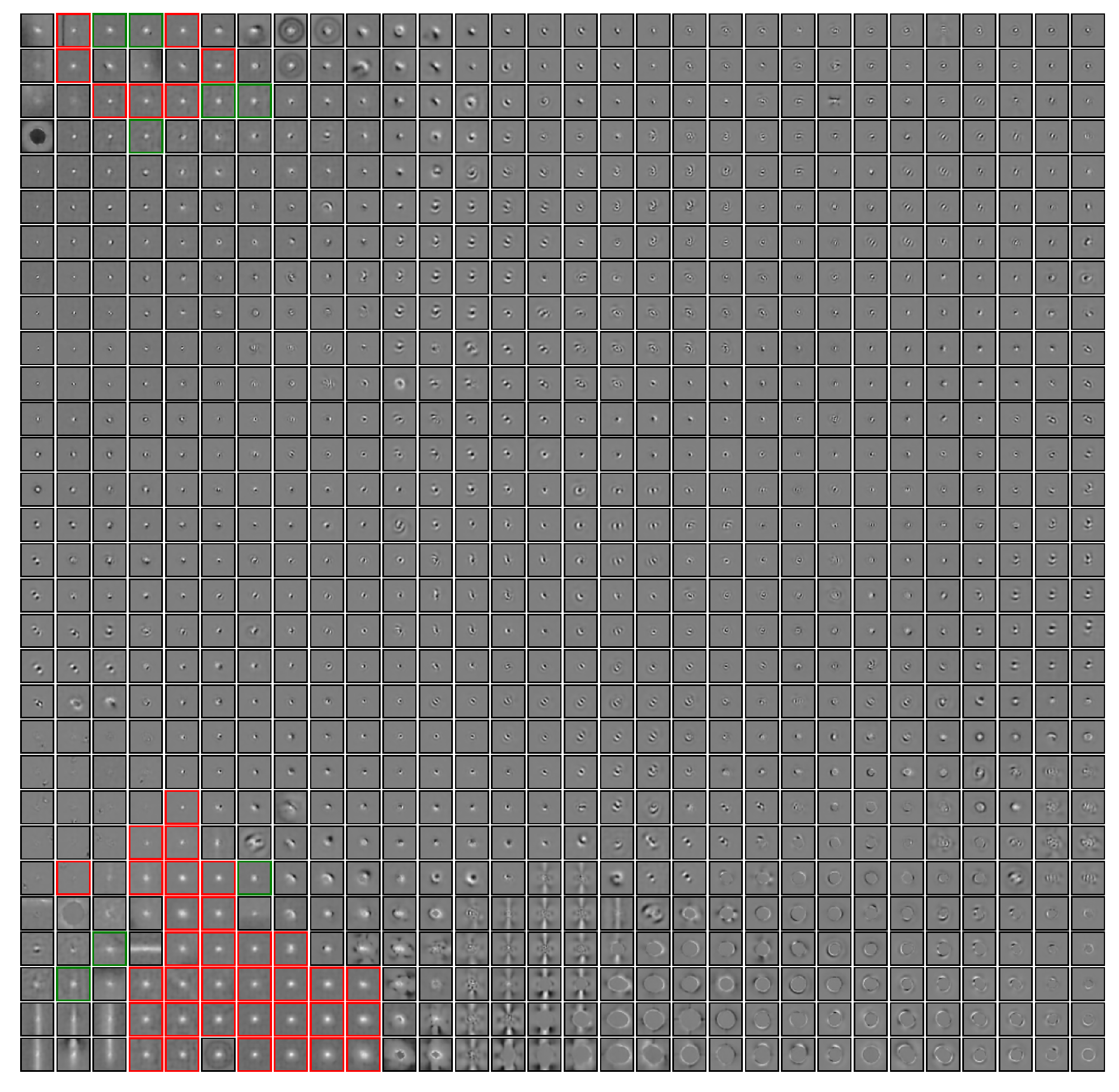}
    \caption{The decoded PVs on the DESOM map of the best model configuration showed in Table \ref{tab:best_desom}. The {\it red} bordered PV indicate the selected PVs in the reference PV selection (see text in \S\ref{sec:model_evaluation}). The {\it green} bordered PVs indicates the PVs which are selected manually to further improve the model performance.}
    \label{fig:best_desom_map}
\end{figure*}

The AEs constructed with different training approaches and complexities share similar averaged reconstruction loss $L_{\rm dec}$ indicating similar performance. The loss can be used to perform model comparison. However, we cannot conclude whether the AE performs well based on its loss value alone. Additionally, we manually compare the reconstructed outputs and the raw inputs to see if the AE performance is reasonable. Figure \ref{fig:off_ae_reconstruct} shows some examples illustrating the AE performance. Since the main objective of the AE in this work is to generate a compressed representation for the input to train the SOM layer, the AE does not need to be perfect but has to be able to pick up enough details from the input for classification purposes. In fact, a more complex AE can improve the image reconstruction, however it also results in a larger latent space which requires a more complex SOM to learn. Therefore, we should always keep our AE simple but just good enough to reproduce the main details of the inputs in order to maximize the efficiency of the SOM training.

Although the AE trained with the DC data set provides a reasonable performance, the performance from the SOM layer is unexpectedly bad. We expect that the subtraction artifacts of a similar type should group together in the latent space. With this assumption, we should be able to conclude the different subtraction issue based on the predicted PVs. However, the predicted PVs generated by the DC-trained DESOM show a completely different morphology from the original inputs (see Figure \ref{fig:off_desom_reconstruct}).

The poor performance of the DC-trained DESOM can be explained by the fact that the SOM prediction is not transformation (rotation and/or translation) invariant \citep{pgi15,tst21}. Thus, two inputs with the same pattern could possibly be separated far apart from each other in the latent space if only the location or the orientation of the pattern is different. As we can see in Figure \ref{fig:off_desom_reconstruct}, while the last three inputs have identical patterns but with offsets, their PVs look completely different, and none of them is a good representation of that pattern. It implies that the DC-trained DESOM fails to group them together in the latent space. Therefore, we are motivated to train our DESOM with the SC data set (see \S\ref{sec:sc_approach}).

We train our DESOM model with the SC data set using different training approaches and find that the separated training approach (\S\ref{subsubsec:separate}) provides the best performance. We perform grid searching on the hyperparameter space of the DESOM. We compare $L_{\rm dec}$ and $L_{\rm som}$ of different model configurations. The best model configuration is shown in Table \ref{tab:best_desom}. The encoder in our best DESOM consists of five hidden layers, three {\tt Conv2D} layers and two {\tt dense} layers. The decoder is symmetric to the encoder built with {\tt dense} layers and {\tt Conv2DTranspose} layers. 

\begin{figure}
\centering
	\includegraphics[width=0.3\textwidth]{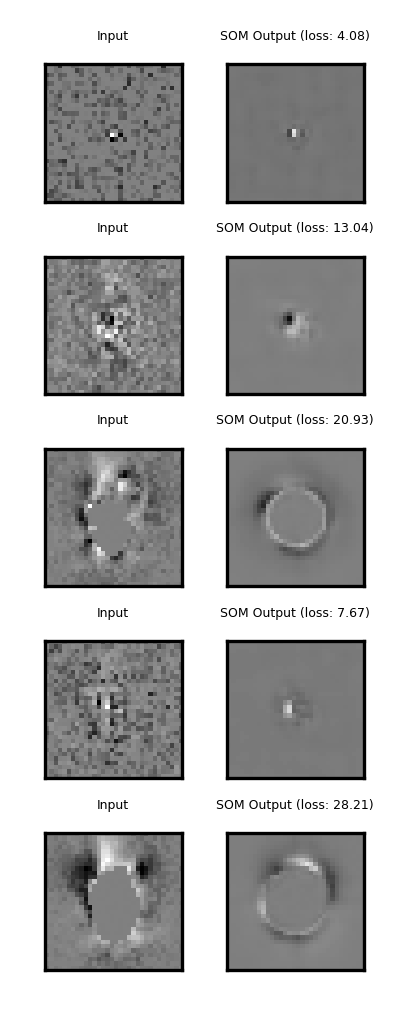}
    \caption{Examples of some cutout inputs and their decoded PVs predicted by the SC-trained DESOM. The SOM loss $L_{\rm som}$ shows a large variation depending on the complexity of the image pattern. These examples show that the DESOM model performs noise reduction and identify the general patterns of the inputs at the same time.}
    \label{fig:on_desom_reconstruct}
\end{figure}

We also show the $30\times30$ DESOM output map in Figure \ref{fig:best_desom_map}. The separation between two PVs on the SOM map is proportional to the Euclidean distance between those corresponding clusters in the latent space. However, we can see that the point-like PVs concentrate at both top left and bottom left corners, which are supposed to stay close with each other. We randomly pick one of the PVs at the top left corner and calculate the Euclidean distance between that PV and those at the bottom right corner. The distance is comparable to the intra-distance of the top left PV clusters. This result indicates that the two corners are very close to each other. The DESOM map shows that it successfully groups the PVs with qualitatively similar patterns together. We can see, in Figure \ref{fig:on_desom_reconstruct}, that the PVs are able to capture the key features from those corresponding inputs.

\begin{table}
	\centering
	\caption{Best DESOM model configuration.}\label{tab:best_desom}
	\begin{tabular}{lr}
		\hline
		Parameter & Value \\
		\hline
		Training set & SC data set \\
		Training approach & Separate-trained \\
		$1^{\rm st}$ hidden layer & 32 neurons ({\tt Conv2D+MaxPooling2D}) \\
		$2^{\rm nd}$ hidden layer & 64 neurons ({\tt Conv2D+MaxPooling2D}) \\
		$3^{\rm rd}$ hidden layer & 128 neurons ({\tt Conv2D+MaxPooling2D}) \\
		$4^{\rm th}$ hidden layer & 512 neurons ({\tt dense}) \\
		$5^{\rm th}$ hidden layer & 120 neurons ({\tt dense}) \\
		Decoder layers & {\tt dense+Conv2DTranspose} \\
		SOM map size & $30\times30$ \\ 
		$T_{\rm max}$ & 10 \\
		$T_{\rm min}$ & 0.01 \\
		Training iterations & 15000 \\
		\hline
	\end{tabular}
\end{table}

\subsection{Model Evaluation}\label{sec:model_evaluation}
Before testing our model accuracy, we visually inspect all DESOM maps generated by the models with different training approaches and hyperparameter configurations. Since most of the DESOM maps generated with the combine-training approach (see \S\ref{sec:combine_train_approach} for more details) have two main issues, some of the PVs are random noise and some other PVs look identical, we decide to deploy the separate-training approach (see \S\ref{subsubsec:separate}) instead. For the rest of this paper, all results are presented based on the separate-training approach.

In order to apply the DESOM model to the real-bogus classification, we need to select which PVs should be classified as real. Since each PV can be labeled as real class or bogus class, there are $2^{900}$ different permutations for a $30\times30$ DESOM map. 

We compare the model performance between DESOM and VGG16 by studying their receiver operating characteristic (ROC) curves. The ROC curve can be generated by gradually moving the decision boundary. However, for the DESOM model, there is no specific order to switch on and off the PVs. With different switching order, $900!$ ROC curves can be generated. To generate some good representatives out of all ROC curves, we decide to use VGG16 predictions as the reference to assign the switching order for the PVs. We apply the DESOM model to our SC test set such that each detection falls onto one of the PVs. Since each test detection comes with the real-bogus score predicted by the VGG16 model, we can order the PVs by the median values of the VGG16 scores. The PV with a higher VGG16 median score means it is more likely to be real. To begin with, we ``switch on'' all PVs. In the other words, we label all PVs to be real at first. We then ``switch off'' the PVs one by one starting from the one with the lowest median VGG score. In the switch-off process, the ROC curve can be generated as the false positive rate (FPR) decreases with the increasing missed detection rate (MDR). Here we use our MP test set to estimate the FPR and the MDR in order to make sure that each test sample has been reviewed manually. 

\begin{figure}
\centering
	\includegraphics[width=0.48\textwidth]{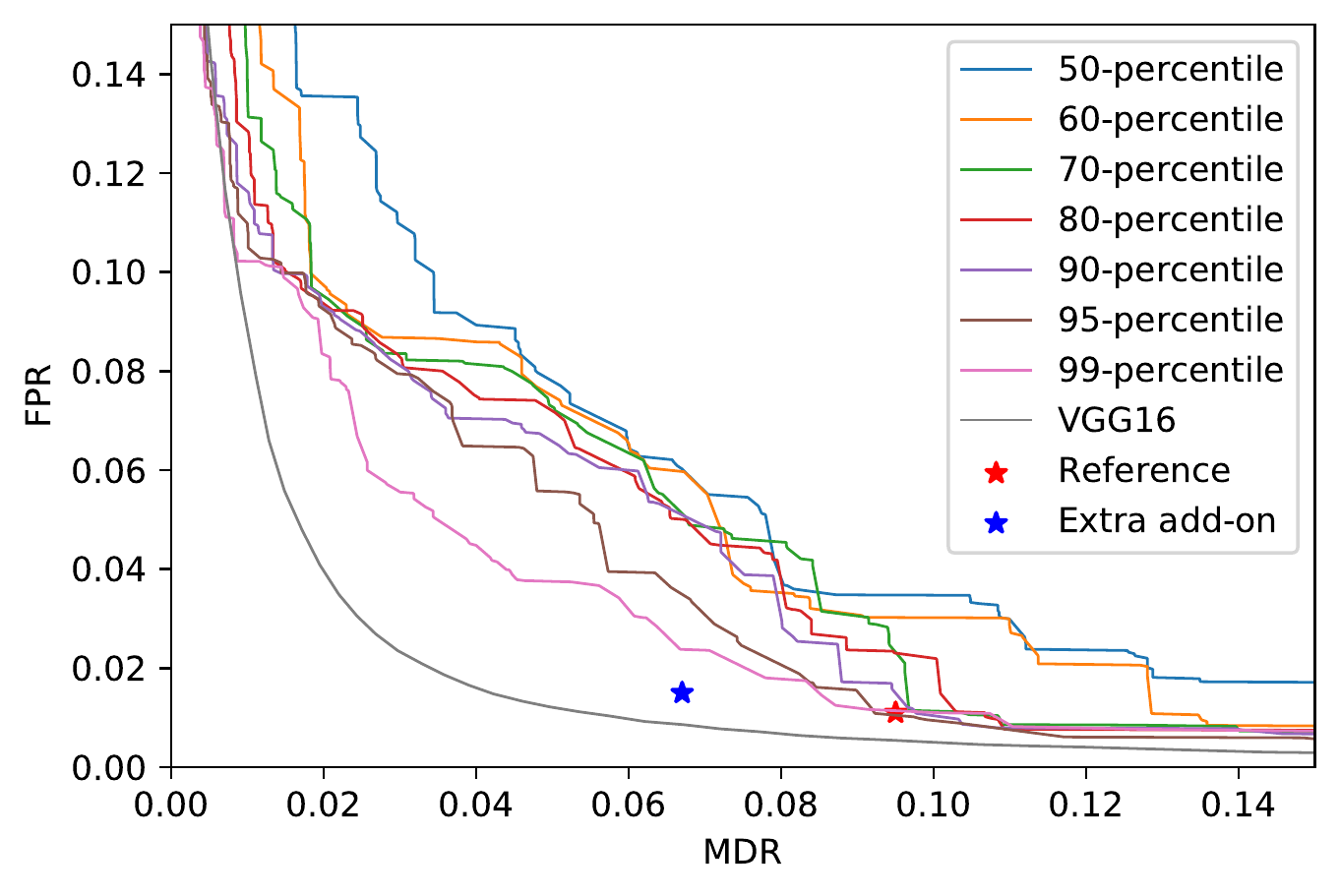}
    \caption{ROC curves generated by ``switching off'' the ordered PVs on the DESOM map one by one. Different colors represent the PVs ordered by different percentile values of the VGG16 score (see text in \S\ref{sec:model_evaluation} for more details). The FoM of the reference PV selection is about $9$--$10\%$. We also plot the ROC curve of the VGG16 classifier. The {\it blue star} indicates the performance of the modified reference PV selection with the inclusion of eight extra PVs (see {\it green bordered} PVs in Figure \ref{fig:best_desom_map}).}
    \label{fig:desom_roc}
\end{figure}

To obtain the best ROC curve, we experimented with switching off the PVs in different orders. We repeat the above procedure with different percentile VGG16 scores instead of the median score. Figure \ref{fig:desom_roc} shows that using 99-percentile VGG16 score to order the PVs generates the best ROC curve. We define the reference PV selection as the one with the lowest MDR at ${\rm FPR}=1\%$. The MDR at ${\rm FPR}=1\%$ is also called the figure of merit (FoM). With these definitions, the FoM of the reference PV selection is about $9$--$10\%$ (see the {\it red star} in Figure \ref{fig:desom_roc}). We also specify the reference PV selection with the {\it red bordered} PVs in Figure \ref{fig:best_desom_map}. By comparing with the ${\rm FoM}\approx6\%$ of the VGG16 model, the reference PV selection performs slightly worse than the VGG16 classifier.

We further modify our reference PV selection by including some extra PV manually. We inspect the selection and find that some of the unselected PVs look real. By including eight extra PVs (indicated by the {\it green bordered} PVs in Figure \ref{fig:best_desom_map}) to the reference PV selection, the MDR drops to $6.6\%$ and the FPR climbs to $1.5\%$ respectively (see the {\it blue star} in Figure \ref{fig:desom_roc}). This exercise demonstrates that further adjustments to the PV selection can improve the performance of the DESOM classifier in combination with the VGG16 classifier.

In practice, since generating ROC curve is not necessary unless for model comparison, we can manually select all PVs which look like genuine detections without any help from other supervised classifiers.

\begin{figure}
\centering
	\includegraphics[width=0.5\textwidth]{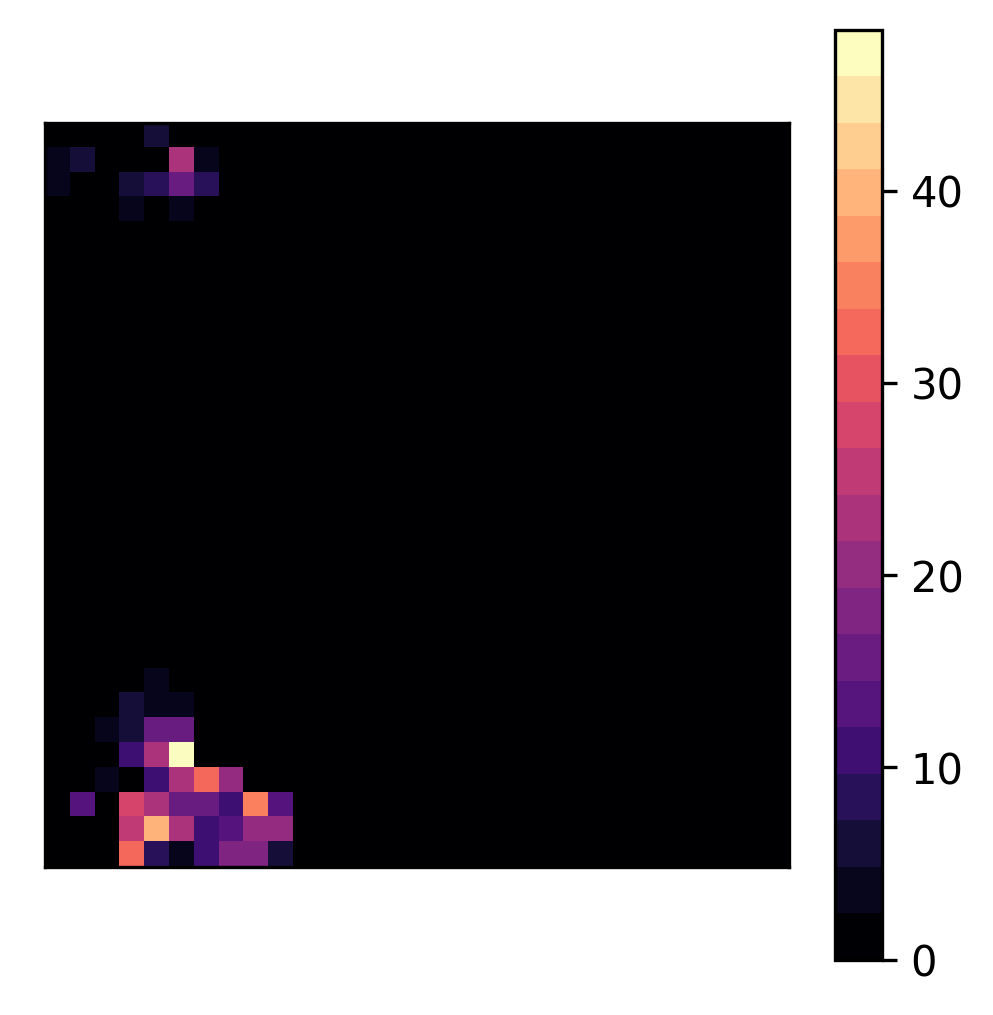}
    \caption{The heatmap presents the performance of the PVs. The color of each cell represents the ratio between the probabilities of a real detection and a bogus detection falling into that PV.}
    \label{fig:real2bogus_ratio}
\end{figure}

In order to visualize the performance of the PVs, for each PV, we calculate the ratio between the probabilities of a real detection and a bogus detection being classified into that PV. The heat map in Figure \ref{fig:real2bogus_ratio} represents the ratio values. We can see that the ratios of those selected PVs are much higher than the others, indicating that a real detection is more likely to fall into those PVs than a bogus detection does.

\section{Discussion} \label{sec:discussion}
We have constructed a deep-embedded self-organizing map (DESOM) model to classify detections on difference images produced with the GOTO prototype instrument, with the objective of improving our ability to distinguish between real and ``bogus'' detections.
We compare our DESOM model with the existing VGG16 classifier by discussing their strengths and weaknesses in \S\ref{sec:desom_vs_vgg}. In \S\ref{sec:usages}, we discuss the potential usages of DESOM beyond real-bogus classification. We also discuss the limitations of the DESOM model in \S\ref{sec:desom_lim}.

\subsection{Comparison Between DESOM and VGG16 Classifier}\label{sec:desom_vs_vgg}
A neural network model is usually considered to be working as a ``black box''. It is very challenging to visually understand the prediction logic. Shifting the decision boundary can improve either the MDR or the FPR. However, balancing the MDR and the FPR does not fundamentally improve the classifier. To do so, we need to re-train the classifier with more data or different model configuration.

On the other hand, DESOM provides users more flexibility in the classification process. Users can simply adjust the model by selecting more or fewer real-labeled PVs on the DESOM map. Since the selection of PVs is done by visual inspection, the adjustment is usually explicable. However, unlike the VGG16 classifier, DESOM is unable to generate a probability score representing how likely a detection is real.

For the VGG16 classifier, we can only shift the entire decision boundary. The step size of the shift can be infinitesimal. However, we cannot control the shape of the decision boundary. In the case of a particular pattern which is always misclassified, the ideal way to improve the performance on that pattern is to only shift the corresponding part of the decision boundary, which cannot be achieved by adjusting the decision boundary of the VGG16 classifier. Unlike the VGG16 classifier, DESOM forms clusters to cover the entire latent space. We can simply change the label of a particular PV if we find that the corresponding PV contains too many false predictions, allowing us to fine-tune the decision boundary. On the other hand, this change is discrete rather than continuous. When we change the label of a PV, it is equivalent to swapping between the true predictions and the false predictions of the PV.

\subsection{Other Potential Usages}\label{sec:usages}
The DESOM model treats real-bogus classification problem as a multi-class classification problem. With the visual inspection of the DESOM map in Figure \ref{fig:best_desom_map}, only $\sim5\%$ of the PVs should be used to represent the clusters of real detections. For the rest of the PVs, they represent the bogus detections with different morphologies and subtraction issues. With this characteristic, we can use DESOM to flag the detections based on their representative shapes.

The DESOM map can also be used to evaluate the performance of other classifiers. Different classifiers usually have different weaknesses. In order to improve the performance of a classifier, we may need to identify which types of patterns the classifier is relatively weak at. To do so, we can study the distribution function of the real-bogus score for each PV cluster. Visual comparison of the detection thumbnails located at both ends of the score distribution can help us understand what causes the confusion of the classifier within the same PV class.

\subsection{Limitations of DESOM} \label{sec:desom_lim}
There are two limitations using DESOM as a real-bogus classifier. First, since DESOM is a self-supervised clustering model, it cannot generate a probability score as a prediction. Second, some of the detections would be lost in the SC source extraction process (see \S\ref{sec:sc_approach}).

With the above two limitations, our DESOM classifier is not ideal to replace the current VGG16 classifier. However, there are two ways of implementing the DESOM output. As we have discussed in \S\ref{sec:usages}, DESOM can be used as a flagging system to provide extra information for each of the detections. Therefore, the original real-bogus score is preserved. Another way of implementing the DESOM model is to build a stack model with the VGG16 classifier. However, a detailed discussion of this approach is outside the scope of this work.

\section{Conclusion} \label{sec:conclusion}
In this work, we demonstrate how to apply a self-supervised learning  approach to the ``real-bogus'' classification problem for difference imaging. The algorithm we used is the deep-embedded self-organizing map (DESOM), a combination of autoencoder (AE) and self-organizing map (SOM) algorithms.

We use $32\times32$ normalized detection thumbnails extracted from the difference images to be the inputs of the DESOM model. We find that using the detection coordinates obtained from the science image to extract the thumbnails can significantly improve the DESOM performance.

We obtain our best DESOM model by training the AE and the SOM layer separately. The figure of merit (FoM) of the DESOM classifier is about $9$--$10\%$. Since the DESOM performance highly depends on the selection of the real prototype vectors (PVs), we show that, by adding a few extra PVs, the DESOM classifier can further be improved (${\rm MDR}=6.6\%$ and ${\rm FPR}=1.5\%$).

The major advantage of the DESOM model is the flexibility of its PV selection, allowing the user fine-control of the shape of the decision boundary. Since the DESOM model treats the real-bogus classification problem as a multi-class  problem, we suggest the best use of DESOM is to build a flagging system for detections, in combination with a probabilistic classification. On top of that, DESOM output map can be used to evaluate the performance of other typical real-bogus classifiers.

\section*{Acknowledgements}
The Gravitational-wave Optical Transient Observer (GOTO) project acknowledges the support of the Monash-Warwick Alliance; Warwick University; Monash University; Sheffield University; University of Leicester; Armagh Observatory \& Planetarium; the National Astronomical Research Institute of Thailand (NARIT); the Instituto de Astrof\'isica de Canarias (IAC) and the University of Turku. RLCS and POB acknowledge support from STFC. RB, MK and DMS acknowledge support from the ERC under the European Union's Horizon 2020 research and innovation programme (grant agreement No. 715051; Spiders). VSD and MJD acknowledge the support of a Leverhulme Trust Research Project Grant. DMS acknowledges support from the Consejer\'ia de Econom\'ia, Conocimiento y Empleo del Gobierno de Canarias and the European Regional Development Fund (ERDF) under grant with reference ProID2020010104.

\section*{Data availability}
{Data files covering the system throughput and some of the software packages are available via public github repositories under \url{https://github.com/GOTO-OBS/}. Prototype data was mainly used for testing and commissioning and a full release of all data is not foreseen. Some data products will be available as part of planned GOTO public data releases.}




\bibliographystyle{mnras}
\bibliography{all} 

\begin{thebibliography}{}
\makeatletter
\relax
\def\mn@urlcharsother{\let\do\@makeother \do\$\do\&\do\#\do\^\do\_\do\%\do\~}
\def\mn@doi{\begingroup\mn@urlcharsother \@ifnextchar [ {\mn@doi@}
  {\mn@doi@[]}}
\def\mn@doi@[#1]#2{\def\@tempa{#1}\ifx\@tempa\@empty \href
  {http://dx.doi.org/#2} {doi:#2}\else \href {http://dx.doi.org/#2} {#1}\fi
  \endgroup}
\def\mn@eprint#1#2{\mn@eprint@#1:#2::\@nil}
\def\mn@eprint@arXiv#1{\href {http://arxiv.org/abs/#1} {{\tt arXiv:#1}}}
\def\mn@eprint@dblp#1{\href {http://dblp.uni-trier.de/rec/bibtex/#1.xml}
  {dblp:#1}}
\def\mn@eprint@#1:#2:#3:#4\@nil{\def\@tempa {#1}\def\@tempb {#2}\def\@tempc
  {#3}\ifx \@tempc \@empty \let \@tempc \@tempb \let \@tempb \@tempa \fi \ifx
  \@tempb \@empty \def\@tempb {arXiv}\fi \@ifundefined
  {mn@eprint@\@tempb}{\@tempb:\@tempc}{\expandafter \expandafter \csname
  mn@eprint@\@tempb\endcsname \expandafter{\@tempc}}}

\bibitem[\protect\citeauthoryear{{Abbott} et~al.,}{{Abbott}
  et~al.}{2016}]{aaa16}
{Abbott} B.~P.,  et~al., 2016, \mn@doi [\apjl] {10.3847/2041-8205/826/1/L13},
  \href {https://ui.adsabs.harvard.edu/abs/2016ApJ...826L..13A} {826, L13}

\bibitem[\protect\citeauthoryear{{Abbott} et~al.,}{{Abbott}
  et~al.}{2017}]{aaa17b}
{Abbott} B.~P.,  et~al., 2017, \mn@doi [\apjl] {10.3847/2041-8213/aa91c9},
  \href {https://ui.adsabs.harvard.edu/abs/2017ApJ...848L..12A} {848, L12}

\bibitem[\protect\citeauthoryear{{Alard} \& {Lupton}}{{Alard} \&
  {Lupton}}{1998}]{al98}
{Alard} C.,  {Lupton} R.~H.,  1998, \mn@doi [\apj] {10.1086/305984}, \href
  {https://ui.adsabs.harvard.edu/abs/1998ApJ...503..325A} {503, 325}

\bibitem[\protect\citeauthoryear{{Andreoni} et~al.,}{{Andreoni}
  et~al.}{2021}]{ack21}
{Andreoni} I.,  et~al., 2021, arXiv e-prints, \href
  {https://ui.adsabs.harvard.edu/abs/2021arXiv210406352A} {p. arXiv:2104.06352}

\bibitem[\protect\citeauthoryear{Baldi}{Baldi}{2012}]{bal12}
Baldi P.,  2012, in Guyon I.,  Dror G.,  Lemaire V.,  Taylor G.,   Silver D.,
  eds,  Proceedings of Machine Learning Research Vol. 27, Proceedings of ICML
  Workshop on Unsupervised and Transfer Learning. PMLR, Bellevue, Washington,
  USA, pp 37--49, \url {https://proceedings.mlr.press/v27/baldi12a.html}

\bibitem[\protect\citeauthoryear{{Bank}, {Koenigstein}  \& {Giryes}}{{Bank}
  et~al.}{2020}]{bkg20}
{Bank} D.,  {Koenigstein} N.,   {Giryes} R.,  2020, arXiv e-prints, \href
  {https://ui.adsabs.harvard.edu/abs/2020arXiv200305991B} {p. arXiv:2003.05991}

\bibitem[\protect\citeauthoryear{{Becker}}{{Becker}}{2015}]{bec15}
{Becker} A.,  2015, {HOTPANTS: High Order Transform of PSF ANd Template
  Subtraction} (\mn@eprint {ascl} {1504.004})

\bibitem[\protect\citeauthoryear{{Berger}}{{Berger}}{2014}]{ber14}
{Berger} E.,  2014, \mn@doi [\araa] {10.1146/annurev-astro-081913-035926},
  \href {https://ui.adsabs.harvard.edu/abs/2014ARA&A..52...43B} {52, 43}

\bibitem[\protect\citeauthoryear{{Berger}, {Fong}  \& {Chornock}}{{Berger}
  et~al.}{2013}]{bfc13}
{Berger} E.,  {Fong} W.,   {Chornock} R.,  2013, \mn@doi [\apjl]
  {10.1088/2041-8205/774/2/L23}, \href
  {https://ui.adsabs.harvard.edu/abs/2013ApJ...774L..23B} {774, L23}

\bibitem[\protect\citeauthoryear{{Berthier}, {Vachier}, {Thuillot}, {Fernique},
  {Ochsenbein}, {Genova}, {Lainey}  \& {Arlot}}{{Berthier}
  et~al.}{2006}]{bvt06}
{Berthier} J.,  {Vachier} F.,  {Thuillot} W.,  {Fernique} P.,  {Ochsenbein} F.,
   {Genova} F.,  {Lainey} V.,   {Arlot} J.~E.,  2006, in {Gabriel} C.,
  {Arviset} C.,  {Ponz} D.,   {Enrique} S.,  eds,  Astronomical Society of the
  Pacific Conference Series Vol. 351, Astronomical Data Analysis Software and
  Systems XV. p.~367

\bibitem[\protect\citeauthoryear{{Bertin} \& {Arnouts}}{{Bertin} \&
  {Arnouts}}{1996}]{ba96}
{Bertin} E.,  {Arnouts} S.,  1996, \mn@doi [\aaps] {10.1051/aas:1996164}, \href
  {https://ui.adsabs.harvard.edu/abs/1996A%26AS..117..393B} {117, 393}

\bibitem[\protect\citeauthoryear{{Blanchard} et~al.,}{{Blanchard}
  et~al.}{2017}]{bbf17}
{Blanchard} P.~K.,  et~al., 2017, \mn@doi [\apjl] {10.3847/2041-8213/aa9055},
  \href {https://ui.adsabs.harvard.edu/abs/2017ApJ...848L..22B} {848, L22}

\bibitem[\protect\citeauthoryear{{Bloom} et~al.,}{{Bloom} et~al.}{2012}]{brn12}
{Bloom} J.~S.,  et~al., 2012, \mn@doi [\pasp] {10.1086/668468}, \href
  {https://ui.adsabs.harvard.edu/abs/2012PASP..124.1175B} {124, 1175}

\bibitem[\protect\citeauthoryear{{Cabrera-Vives}, {Reyes}, {Förster},
  {Estévez}  \& {Maureira}}{{Cabrera-Vives} et~al.}{2016}]{cfe16}
{Cabrera-Vives} G.,  {Reyes} I.,  {Förster} F.,  {Estévez} P.~A.,
  {Maureira} J.,  2016, in 2016 International Joint Conference on Neural
  Networks (IJCNN). pp 251--258

\bibitem[\protect\citeauthoryear{{Cabrera-Vives}, {Reyes}, {F{\"o}rster},
  {Est{\'e}vez}  \& {Maureira}}{{Cabrera-Vives} et~al.}{2017}]{crf17}
{Cabrera-Vives} G.,  {Reyes} I.,  {F{\"o}rster} F.,  {Est{\'e}vez} P.~A.,
  {Maureira} J.-C.,  2017, \mn@doi [\apj] {10.3847/1538-4357/836/1/97}, \href
  {https://ui.adsabs.harvard.edu/abs/2017ApJ...836...97C} {836, 97}

\bibitem[\protect\citeauthoryear{{Cenko} et~al.,}{{Cenko} et~al.}{2015}]{cup15}
{Cenko} S.~B.,  et~al., 2015, \mn@doi [\apjl] {10.1088/2041-8205/803/2/L24},
  \href {https://ui.adsabs.harvard.edu/abs/2015ApJ...803L..24C} {803, L24}

\bibitem[\protect\citeauthoryear{{Chornock} et~al.,}{{Chornock}
  et~al.}{2017}]{cbk17}
{Chornock} R.,  et~al., 2017, \mn@doi [\apjl] {10.3847/2041-8213/aa905c}, \href
  {https://ui.adsabs.harvard.edu/abs/2017ApJ...848L..19C} {848, L19}

\bibitem[\protect\citeauthoryear{{Coughlin} et~al.,}{{Coughlin}
  et~al.}{2020}]{caa20}
{Coughlin} M.,  et~al., 2020, GRB Coordinates Network, \href
  {https://ui.adsabs.harvard.edu/abs/2020GCN.28841....1C} {28841, 1}

\bibitem[\protect\citeauthoryear{{Coulter} et~al.,}{{Coulter}
  et~al.}{2017}]{cfk17}
{Coulter} D.~A.,  et~al., 2017, \mn@doi [Science] {10.1126/science.aap9811},
  \href {https://ui.adsabs.harvard.edu/abs/2017Sci...358.1556C} {358, 1556}

\bibitem[\protect\citeauthoryear{{Cowperthwaite} et~al.,}{{Cowperthwaite}
  et~al.}{2017}]{cbv17}
{Cowperthwaite} P.~S.,  et~al., 2017, \mn@doi [\apjl]
  {10.3847/2041-8213/aa8fc7}, \href
  {https://ui.adsabs.harvard.edu/abs/2017ApJ...848L..17C} {848, L17}

\bibitem[\protect\citeauthoryear{Daniel}{Daniel}{2013}]{dan13}
Daniel G.~G.,  2013, Artificial Neural Network.
Springer Netherlands, Dordrecht, pp 143--143,
  \mn@doi{10.1007/978-1-4020-8265-8_200980}, \url
  {https://doi.org/10.1007/978-1-4020-8265-8_200980}

\bibitem[\protect\citeauthoryear{{Duev} et~al.,}{{Duev} et~al.}{2019}]{dmm19}
{Duev} D.~A.,  et~al., 2019, \mn@doi [\mnras] {10.1093/mnras/stz2357}, \href
  {https://ui.adsabs.harvard.edu/abs/2019MNRAS.489.3582D} {489, 3582}

\bibitem[\protect\citeauthoryear{{Dyer} et~al.,}{{Dyer} et~al.}{2020}]{dsg20}
{Dyer} M.~J.,  et~al., 2020, in Society of Photo-Optical Instrumentation
  Engineers (SPIE) Conference Series. p. 114457G (\mn@eprint {arXiv}
  {2012.02685}), \mn@doi{10.1117/12.2561008}

\bibitem[\protect\citeauthoryear{Forest, Lebbah, Azzag  \& Lacaille}{Forest
  et~al.}{2019}]{fla19}
Forest F.,  Lebbah M.,  Azzag H.,   Lacaille J.,  2019.

\bibitem[\protect\citeauthoryear{{Gieseke} et~al.,}{{Gieseke}
  et~al.}{2017}]{gbv17}
{Gieseke} F.,  et~al., 2017, \mn@doi [\mnras] {10.1093/mnras/stx2161}, \href
  {https://ui.adsabs.harvard.edu/abs/2017MNRAS.472.3101G} {472, 3101}

\bibitem[\protect\citeauthoryear{{Goldstein} et~al.,}{{Goldstein}
  et~al.}{2017}]{gvb17}
{Goldstein} A.,  et~al., 2017, \mn@doi [\apjl] {10.3847/2041-8213/aa8f41},
  \href {https://ui.adsabs.harvard.edu/abs/2017ApJ...848L..14G} {848, L14}

\bibitem[\protect\citeauthoryear{{Gompertz} et~al.,}{{Gompertz}
  et~al.}{2020}]{gcs20}
{Gompertz} B.~P.,  et~al., 2020, \mn@doi [\mnras] {10.1093/mnras/staa1845},
  \href {https://ui.adsabs.harvard.edu/abs/2020MNRAS.497..726G} {497, 726}

\bibitem[\protect\citeauthoryear{{Hallinan} et~al.,}{{Hallinan}
  et~al.}{2017}]{hcm17}
{Hallinan} G.,  et~al., 2017, \mn@doi [Science] {10.1126/science.aap9855},
  \href {https://ui.adsabs.harvard.edu/abs/2017Sci...358.1579H} {358, 1579}

\bibitem[\protect\citeauthoryear{{Ho} et~al.,}{{Ho} et~al.}{2020}]{hpb20}
{Ho} A. Y.~Q.,  et~al., 2020, \mn@doi [\apj] {10.3847/1538-4357/abc34d}, \href
  {https://ui.adsabs.harvard.edu/abs/2020ApJ...905...98H} {905, 98}

\bibitem[\protect\citeauthoryear{{Jin} et~al.,}{{Jin} et~al.}{2013}]{jcd13}
{Jin} Z.-P.,  et~al., 2013, \mn@doi [\apj] {10.1088/0004-637X/774/2/114}, \href
  {https://ui.adsabs.harvard.edu/abs/2013ApJ...774..114J} {774, 114}

\bibitem[\protect\citeauthoryear{{Kasliwal}, {Korobkin}, {Lau}, {Wollaeger}  \&
  {Fryer}}{{Kasliwal} et~al.}{2017}]{kkl17}
{Kasliwal} M.~M.,  {Korobkin} O.,  {Lau} R.~M.,  {Wollaeger} R.,   {Fryer}
  C.~L.,  2017, \mn@doi [\apjl] {10.3847/2041-8213/aa799d}, \href
  {https://ui.adsabs.harvard.edu/abs/2017ApJ...843L..34K} {843, L34}

\bibitem[\protect\citeauthoryear{{Killestein} et~al.,}{{Killestein}
  et~al.}{2021}]{kls21}
{Killestein} T.~L.,  et~al., 2021, \mn@doi [\mnras] {10.1093/mnras/stab633},
  \href {https://ui.adsabs.harvard.edu/abs/2021MNRAS.503.4838K} {503, 4838}

\bibitem[\protect\citeauthoryear{Kohonen}{Kohonen}{1990}]{koh90}
Kohonen T.,  1990, \mn@doi [Proceedings of the IEEE] {10.1109/5.58325}, 78,
  1464

\bibitem[\protect\citeauthoryear{Kohonen}{Kohonen}{2001}]{koh01}
Kohonen T.,  2001, Self-organizing maps, 3rd edn.
Springer series in information sciences, 30, Springer, \url
  {http://www.worldcat.org/isbn/3540679219}

\bibitem[\protect\citeauthoryear{{Kumar} \& {Zhang}}{{Kumar} \&
  {Zhang}}{2015}]{kz15}
{Kumar} P.,  {Zhang} B.,  2015, \mn@doi [\physrep]
  {10.1016/j.physrep.2014.09.008}, \href
  {https://ui.adsabs.harvard.edu/abs/2015PhR...561....1K} {561, 1}

\bibitem[\protect\citeauthoryear{{Lamb} et~al.,}{{Lamb} et~al.}{2019}]{ltl19}
{Lamb} G.~P.,  et~al., 2019, \mn@doi [\apj] {10.3847/1538-4357/ab38bb}, \href
  {https://ui.adsabs.harvard.edu/abs/2019ApJ...883...48L} {883, 48}

\bibitem[\protect\citeauthoryear{{Li} et~al.,}{{Li} et~al.}{2012}]{llt12}
{Li} L.,  et~al., 2012, \mn@doi [\apj] {10.1088/0004-637X/758/1/27}, \href
  {https://ui.adsabs.harvard.edu/abs/2012ApJ...758...27L} {758, 27}

\bibitem[\protect\citeauthoryear{{Margutti} et~al.,}{{Margutti}
  et~al.}{2017}]{mbf17}
{Margutti} R.,  et~al., 2017, \mn@doi [\apjl] {10.3847/2041-8213/aa9057}, \href
  {https://ui.adsabs.harvard.edu/abs/2017ApJ...848L..20M} {848, L20}

\bibitem[\protect\citeauthoryear{{Mong} et~al.,}{{Mong} et~al.}{2020}]{mag20}
{Mong} Y.~L.,  et~al., 2020, \mn@doi [\mnras] {10.1093/mnras/staa3096}, \href
  {https://ui.adsabs.harvard.edu/abs/2020MNRAS.499.6009M} {499, 6009}

\bibitem[\protect\citeauthoryear{{Mong} et~al.,}{{Mong} et~al.}{2021}]{mag21}
{Mong} Y.~L.,  et~al., 2021, \mn@doi [\mnras] {10.1093/mnras/stab2499}, \href
  {https://ui.adsabs.harvard.edu/abs/2021MNRAS.507.5463M} {507, 5463}

\bibitem[\protect\citeauthoryear{Munro}{Munro}{2010}]{mun10}
Munro P.,  2010, Backpropagation.
Springer US, Boston, MA, pp 73--73, \mn@doi{10.1007/978-0-387-30164-8_51}, \url
  {https://doi.org/10.1007/978-0-387-30164-8_51}

\bibitem[\protect\citeauthoryear{{O'Shea} \& {Nash}}{{O'Shea} \&
  {Nash}}{2015}]{on15}
{O'Shea} K.,  {Nash} R.,  2015, arXiv e-prints, \href
  {https://ui.adsabs.harvard.edu/abs/2015arXiv151108458O} {p. arXiv:1511.08458}

\bibitem[\protect\citeauthoryear{{Piran}}{{Piran}}{2004}]{piran04}
{Piran} T.,  2004, \mn@doi [Reviews of Modern Physics]
  {10.1103/RevModPhys.76.1143}, \href
  {https://ui.adsabs.harvard.edu/abs/2004RvMP...76.1143P} {76, 1143}

\bibitem[\protect\citeauthoryear{{Polsterer}, {Gieseke}  \& {Igel}}{{Polsterer}
  et~al.}{2015}]{pgi15}
{Polsterer} K.~L.,  {Gieseke} F.,   {Igel} C.,  2015, in {Taylor} A.~R.,
  {Rosolowsky} E.,  eds,  Astronomical Society of the Pacific Conference Series
  Vol. 495, Astronomical Data Analysis Software an Systems XXIV (ADASS XXIV).
  p.~81

\bibitem[\protect\citeauthoryear{{Rau} et~al.,}{{Rau} et~al.}{2009}]{rkl09}
{Rau} A.,  et~al., 2009, \mn@doi [\pasp] {10.1086/605911}, \href
  {https://ui.adsabs.harvard.edu/abs/2009PASP..121.1334R} {121, 1334}

\bibitem[\protect\citeauthoryear{{Savchenko} et~al.,}{{Savchenko}
  et~al.}{2017}]{sfk17}
{Savchenko} V.,  et~al., 2017, \mn@doi [\apjl] {10.3847/2041-8213/aa8f94},
  \href
  {https://ui-adsabs-harvard-edu.ezproxy.lib.monash.edu.au/abs/2017ApJ...848L..15S}
  {848, L15}

\bibitem[\protect\citeauthoryear{{Simonyan} \& {Zisserman}}{{Simonyan} \&
  {Zisserman}}{2014}]{sz14}
{Simonyan} K.,  {Zisserman} A.,  2014, arXiv e-prints, \href
  {https://ui.adsabs.harvard.edu/abs/2014arXiv1409.1556S} {p. arXiv:1409.1556}

\bibitem[\protect\citeauthoryear{{Smith} et~al.,}{{Smith} et~al.}{2020}]{ssy20}
{Smith} K.~W.,  et~al., 2020, \mn@doi [\pasp] {10.1088/1538-3873/ab936e}, \href
  {https://ui.adsabs.harvard.edu/abs/2020PASP..132h5002S} {132, 085002}

\bibitem[\protect\citeauthoryear{{Steeghs} et~al.,}{{Steeghs}
  et~al.}{2021}]{sga21}
{Steeghs} D.,  et~al., 2021, arXiv e-prints, \href
  {https://ui.adsabs.harvard.edu/abs/2021arXiv211005539S} {p. arXiv:2110.05539}

\bibitem[\protect\citeauthoryear{{Tanvir}, {Levan}, {Fruchter}, {Hjorth},
  {Hounsell}, {Wiersema}  \& {Tunnicliffe}}{{Tanvir} et~al.}{2013}]{tlf13}
{Tanvir} N.~R.,  {Levan} A.~J.,  {Fruchter} A.~S.,  {Hjorth} J.,  {Hounsell}
  R.~A.,  {Wiersema} K.,   {Tunnicliffe} R.~L.,  2013, \mn@doi [\nat]
  {10.1038/nature12505}, \href
  {https://ui.adsabs.harvard.edu/abs/2013Natur.500..547T} {500, 547}

\bibitem[\protect\citeauthoryear{{Teimoorinia}, {Shishehchi}, {Tazwar}, {Lin},
  {Archinuk}, {Gwyn}  \& {Kavelaars}}{{Teimoorinia} et~al.}{2021}]{tst21}
{Teimoorinia} H.,  {Shishehchi} S.,  {Tazwar} A.,  {Lin} P.,  {Archinuk} F.,
  {Gwyn} S. D.~J.,   {Kavelaars} J.~J.,  2021, \mn@doi [\aj]
  {10.3847/1538-3881/abea7e}, \href
  {https://ui.adsabs.harvard.edu/abs/2021AJ....161..227T} {161, 227}

\bibitem[\protect\citeauthoryear{Wang, Huang, Wang  \& Wang}{Wang
  et~al.}{2014}]{whw14}
Wang W.,  Huang Y.,  Wang Y.,   Wang L.,  2014, in 2014 IEEE Conference on
  Computer Vision and Pattern Recognition Workshops. pp 496--503,
  \mn@doi{10.1109/CVPRW.2014.79}

\bibitem[\protect\citeauthoryear{{Zhang}, {Fan}, {Dyks}, {Kobayashi},
  {M{\'e}sz{\'a}ros}, {Burrows}, {Nousek}  \& {Gehrels}}{{Zhang}
  et~al.}{2006}]{zfd06}
{Zhang} B.,  {Fan} Y.~Z.,  {Dyks} J.,  {Kobayashi} S.,  {M{\'e}sz{\'a}ros} P.,
  {Burrows} D.~N.,  {Nousek} J.~A.,   {Gehrels} N.,  2006, \mn@doi [\apj]
  {10.1086/500723}, \href
  {https://ui.adsabs.harvard.edu/abs/2006ApJ...642..354Z} {642, 354}

\makeatother
\end{thebibliography}








\end{document}